\documentclass[11pt,a4paper]{article}
\usepackage{times,latexsym}
\usepackage{url}
\usepackage[T1]{fontenc}

\usepackage{graphicx}
\usepackage{multicol}
\usepackage{multirow}
\usepackage[most]{tcolorbox}

\usepackage[acceptedWithA]{tacl2021v1}
\usepackage{xspace,mfirstuc,tabulary}

\usepackage{array}
\usepackage[table]{xcolor}
\usepackage{tcolorbox}
\usepackage{booktabs}
\usepackage{algorithm}
\usepackage{algorithmic}
\usepackage{xspace} 
\usepackage{booktabs}
\usepackage{amsmath}
\usepackage{newfloat}
\usepackage{listings}
\usepackage{colortbl}
\usepackage[normalem]{ulem}
\usepackage{fontawesome5}
\useunder{\uline}{\ul}{}

\definecolor{step1}{HTML}{FFCCCC} % light red
\definecolor{step2}{HTML}{CCFFCC} % light green
\definecolor{step3}{HTML}{CCCCFF} % light blue
\definecolor{answercol}{HTML}{FFF2CC} % light yellow for answer
\definecolor{headerbg}{HTML}{005F73}
\definecolor{textbg}{HTML}{F9F9F9}

\definecolor{step1text}{HTML}{CC0000}    % dark red
\definecolor{step2text}{HTML}{006600}    % dark green
\definecolor{step3text}{HTML}{000099}    % dark blue
\definecolor{answertext}{HTML}{996600}   % dark gold/brown

\definecolor{answercolor}{HTML}{996600}     % brown for answers
\definecolor{colcolor}{HTML}{000099}        % blue for relevant columns
\definecolor{rowcolor}{HTML}{CC0000}  
\definecolor{correct}{HTML}{A6ECA8}   % Light green
\definecolor{incorrect}{HTML}{F28B82} % Light red

\newif\iftaclinstructions
\taclinstructionsfalse % AUTHORS: do NOT set this to true
\iftaclinstructions
\renewcommand{\confidential}{}
\renewcommand{\anonsubtext}{(No author info supplied here, for consistency with
TACL-submission anonymization requirements)}
\newcommand{\instr}
\fi

\iftaclpubformat % this "if" is set by the choice of options

\else

\fi

\newcounter{promptctr}
\renewcommand{\thepromptctr}{Prompt~\Alph{promptctr}}

\definecolor{examplebg}{HTML}{D1FFBD}
\tcbset{
  bw:domain/.style={
    colback=white,
    colframe=black!40,
    coltitle=black,
    fonttitle=\bfseries\sffamily,   % ✅ fixed
    colbacktitle=examplebg,
    boxrule=0.5pt,
    titlerule=0pt,
    arc=2pt,
    left=4pt,right=4pt,top=4pt,bottom=4pt,
    toptitle=2pt,bottomtitle=2pt,
  }
}

\newcommand{\mytcbinput}[5]{% file, title, num, style, label
  \refstepcounter{promptctr}% increment + make it ref'able
  \phantomsection              % create hyperlink anchor
  \begin{tcolorbox}[title={\thepromptctr: #2},#4]
    \lstinputlisting{#1}
    \label{#5} % assign label for referencing
  \end{tcolorbox}%
}

\newcommand{\mytcbinputwide}[5]{% file, title, num, style, label
  \begin{figure*}[t]
  \centering
  \refstepcounter{promptctr}
  \phantomsection
  \begin{tcolorbox}[title={\thepromptctr: #2},#4,width=\textwidth,enhanced]
    \lstinputlisting{#1}
    \label{#5}
  \end{tcolorbox}
  \vspace{-4pt}
  \end{figure*}
}

\date{}
\newcommand{\methodName}{{\sc TraceBack}\xspace}

\newcommand{\metricName}{{\sc FAIRScore}\xspace}

\newcommand{\benchmarkName}{{\sc CITEBench}\xspace}

\newcommand{\promptref}[1]{%
  \hyperref[#1]{\ref*{#1}}%
}

\newcommand{\promptrefp}[1]{%
  \hyperref[#1]{\ref*{#1} (p.~\pageref*{#1})}%
}

\title{\methodName: Multi-Agent Decomposition for Fine-Grained \\ Table Attribution}

\author{%
  \raisebox{0.75ex}{\includegraphics[height=2ex]{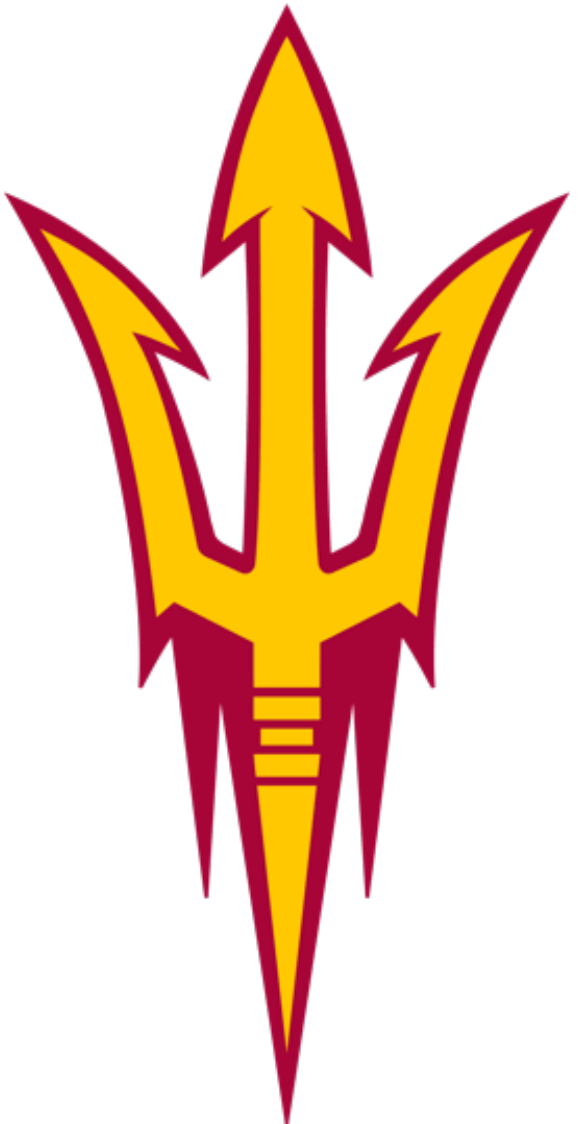}}Tejas Anvekar\Thanks{ \em contributed equally} \quad
  \raisebox{0.75ex}{\includegraphics[height=2ex]{./images/asu_logo.pdf}}Junha Park\footnotemark[1] \quad
  \raisebox{0.75ex}{\includegraphics[height=2ex]{./images/asu_logo.pdf}}Rajat Jha\footnotemark[1] \quad
  \raisebox{0.75ex}{\includegraphics[height=2ex]{./images/asu_logo.pdf}}Devanshu Gupta \quad
  \raisebox{0.75ex}{\includegraphics[height=2ex]{./images/asu_logo.pdf}}Poojah Ganesan \\
  \textbf{\raisebox{0.75ex}{\includegraphics[height=1.75ex]{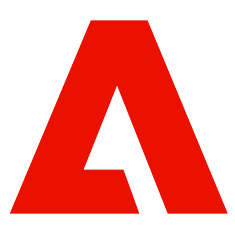}}Puneeth Mathur} \quad
  \textbf{\raisebox{0.75ex}{\includegraphics[height=2ex]{./images/asu_logo.pdf}}Vivek Gupta} \\
  \raisebox{0.75ex}{\includegraphics[height=2ex]{./images/asu_logo.pdf}}\hspace{0.25ex}Arizona State University \quad
  \raisebox{0.75ex}{\includegraphics[height=1.75ex]{./images/adobe_logo.pdf}}\hspace{0.25ex}Adobe Research \\[0.4em]
  \faGlobe\;\href{https://coral-lab-asu.github.io/TraceBack/}{Project-Page} \quad \faGithub\;\href{https://github.com/CoRAL-ASU/TraceBack}{Code}\\
 \texttt{\{tanvekar, jpark284, rjha16, dgupta77, pganesa4, vgupt140\}@asu.edu}, \\ \texttt{puneetm@adobe.com} 
}
\begin{document}
\maketitle
% \begin{abstract}
% Accessing answers from structured tables requires not only accurate generation but also trust in the underlying reasoning. However, most Question Answering (QA) systems over tables lack fine-grained attribution, often producing correct answers without revealing which cells support them. This undermines transparency and limits trust, especially in high-stakes domains. To address this, we introduce \methodName, a modular, multi-agent framework that performs scalable cell-level attribution for single-table QA by pruning tables to identify relevant rows and columns, decomposing questions, and mapping answer spans to specific table cells. By decomposing questions into semantically coherent sub-questions, our method surfaces intermediate reasoning steps and captures implicit cells often missed by existing attribution methods. To support robust evaluation, we release \benchmarkName, benchmark spanning multiple tabular QA dataset such as ToTTo, FetaQA, and AITQA, with gold-labeled cell alignments. We further propose \metricName, a reference-less metric that compares atomic facts derived from predicted cells and answers to compute attribution precision and recall without human labels. \methodName outperforms various evaluation benchmarks and advances interpretability and trust in table-based QA.
% \end{abstract}
\begin{abstract}
Question answering (QA) over structured tables requires not only accurate answers but also transparency about which cells support them.  Existing table QA systems rarely provide fine-grained attribution, so even correct answers often lack verifiable grounding, limiting trust in high-stakes settings. We address this with \methodName\, a modular multi-agent framework for scalable, cell-level attribution in single-table QA. \methodName\ prunes tables to relevant rows and columns, decomposes questions into semantically coherent sub-questions, and aligns each answer span with its supporting cells, capturing both explicit and implicit evidence used in intermediate reasoning steps. To enable systematic evaluation, we release \benchmarkName, a benchmark with phrase-to-cell annotations drawn from ToTTo, FetaQA, and AITQA. We further propose \metricName, a reference-less metric that compares atomic facts derived from predicted cells and answers to estimate attribution precision and recall without human cell labels. Experiments show that \methodName\ substantially outperforms strong baselines across datasets and granularities, while \metricName\ closely tracks human judgments and preserves relative method rankings, supporting interpretable and scalable evaluation of table-based QA.
\end{abstract}

\section{Introduction}

Question answering (QA) is a core task in NLP with applications ranging from customer support to scientific analysis. 
Large language models (LLMs) have dramatically improved QA quality \cite{brown2020languagemodelsfewshotlearners, achiam2023gpt4}, but they frequently hallucinate or produce factually incorrect statements \cite{ji2023survey}, undermining reliability and user trust \cite{xu2024hallucination, snyder2023early}. 
Even when answers are correct, the underlying reasoning may be spurious or opaque, motivating methods that make both answers and their evidence more transparent.

\begin{figure}[t]
\centering
\resizebox{\linewidth}{!}{%
\begin{tabular}{>{\centering\arraybackslash}p{2.0cm} >{\centering\arraybackslash}p{1.2cm} >{\centering\arraybackslash}p{1.5cm} >{\centering\arraybackslash}p{1.5cm}}
\rowcolor{headerbg}
\color{white}\textbf{Source} & \color{white}\textbf{Cost} & \color{white}\textbf{Efficiency} & \color{white}\textbf{Scalability} \\
Solar Power & \cellcolor{step1}$30$--$50$ & 15--20 & \cellcolor{step2}4 \\
\cellcolor{answercol}{\textcolor{answertext}{Wind Power}}  & \cellcolor{step1}$20$--$40$ & \cellcolor{step3}\textcolor{answertext}{30--45} & \cellcolor{step2}5 \\
Hydropower  & $40$--$70$ & 70--90 & 3 \\
Geothermal  & $50$--$80$ & 90+ & 2 \\
\end{tabular}}

\tcbset{colback=textbg, colframe=white, boxsep=2pt, left=2pt, right=2pt, top=2pt, bottom=2pt, boxrule=0pt, arc=2pt}
\begin{tcolorbox}
\textbf{Question:} Among renewable sources costing $\leq 50$/MWh and scalability $\geq 3$, which is most efficient, and what is its efficiency? \\[1pt]

\textbf{Reasoning:} 
\textcolor{red}{Step 1: Cost Filter} $\rightarrow$ Keep Solar, Wind.  
\textcolor{green!60!black}{Step 2: Scalability Filter} $\rightarrow$ Keep Solar, Wind.  
\textcolor{blue!70!black}{Step 3: Efficiency Selection} $\rightarrow$ Choose Wind (30--45\%).\\

\resizebox{\linewidth}{!}{%
\begin{tabular}{l l l}
\cellcolor{step1}\phantom{X} S1: Cost $\leq 50$ &
\cellcolor{step2}\phantom{X} S2: Scalability $\geq 3$ &
\cellcolor{step3}\phantom{X} S3: Max Efficiency
\end{tabular}
}
\\

\vspace{2pt} \textbf{Answer:} \textcolor{answertext}{Wind Power}, \textcolor{answertext}{30--45\%} efficiency.
\end{tcolorbox}
\caption{Example of fine-grained table attribution: intermediate filters (cost, scalability) and final selection (efficiency) each correspond to specific cells.}
\label{fig:introduction}
\vspace{-1.0em}
\end{figure}

Attribution has therefore become central to evidence-based reasoning in LLMs, linking answers to verifiable sources \cite{gao-etal-2023-rarr, li2023survey}. 
Most prior work focuses on unstructured text and relatively coarse citation; structured data such as tables introduce additional challenges due to schema constraints, hierarchical organization, and multi-step filtering (Figure~\ref{fig:introduction}).

For tabular QA, accurately identifying which cells support the answer is crucial for trust and interpretability. 
However, cell-level attribution remains largely unexplored. 
To date, the only prior work explicitly targeting table attribution, MATSA \cite{mathur2024matsa}, operates at row-column granularity, lacks phrase-level alignment, and focuses solely on final answers. 
It does not expose intermediate evidence, such as start and end time cells used to compute a duration, even though these are essential steps in the reasoning process.

This raises a central question: 
\emph{How can we design an interpretable, scalable framework that traces both answers and intermediate reasoning steps back to precise, cell-level evidence in structured tables?}

To address this, we introduce \benchmarkName, a benchmark for fine-grained attribution in table QA comprising 1{,}500 manually annotated examples from ToTTo \cite{parikh2020totto}, FetaQA \cite{nan2021fetaqa}, and AITQA \cite{katsis2022aitqa}. 
Each example includes cell-level attributions for both final answer spans and intermediate reasoning, enabling supervision for multi-hop and implicit inference.

Building on this resource, we propose \methodName, a modular multi-agent LLM-based framework for post-hoc cell-level attribution over single tables. 
\methodName\ sequentially (i) identifies relevant columns, (ii) filters rows via schema-aware conditions, (iii) decomposes questions into sub-questions aligned with intermediate reasoning steps, and (iv) grounds each sub-answer in specific cells, which are then aligned with answer spans (\autoref{fig:introduction}). 
On the 1{,}500 annotated examples, \methodName\ yields substantial gains over strong baselines in both fine-grained attribution and coverage of intermediate evidence.

Manual cell-level annotation is, however, expensive and subjective, making large-scale evaluation difficult. 
We therefore introduce \metricName, a reference-less metric that compares atomic facts extracted from predicted cells with those in the answer. 
By estimating attribution precision and recall via fact alignment, \metricName\ enables scalable, consistent, and interpretable evaluation across datasets, including unlabeled settings.

Our main contributions are:

\begin{enumerate}
\item We present \benchmarkName, a benchmark of manually annotated QA examples from ToTTo, FetaQA, and AITQA with cell-level attributions covering both final answers and intermediate reasoning steps.
\item We propose \methodName, a multi-agent LLM-based framework for cell-level attribution that combines schema-aware pruning, question decomposition, and fine-grained cell grounding.
\item We develop \metricName, a reference-less evaluation metric based on atomic fact alignment that estimates attribution precision and recall without human cell labels.
\item We conduct extensive experiments and ablations, showing that \methodName\ consistently outperforms strong baselines while \metricName\ tracks human judgments and preserves relative method rankings.
\end{enumerate}

\section{Related Work}
Attribution in QA is central to improving the reliability, interpretability, and trustworthiness of large language models (LLMs), particularly in structured and multi-source settings. Early work, such as FEVER \cite{thorne2018shared}, framed fact verification as attributing claims to supporting evidence, while \cite{evans2021role} emphasized its role in truthful AI. Fake news detection has also been cast as an attribution task \cite{pomerleau2017fake}. In citation evaluation, \cite{gao2023alce} introduced ALCE, a benchmark assessing citation quality alongside answer fluency and correctness. Its reliance on MAUVE \cite{pillutla2021mauve} for fluency scoring can yield unstable results due to sensitivity to output length and style, and correctness remains difficult to automate for multi-step answers.

For structured data, \cite{mathur2024matsa} proposed MATSA, a multi-agent framework for table attribution with an 8.5K QA-pair benchmark from ToTTo \cite{parikh2020totto}, FetaQA \cite{nan2021fetaqa}, and AITQA \cite{katsis2022aitqa}. While effective on short contexts, MATSA is limited to row-column attribution and omits cell-level or long-context QA. Multi-source attribution typically follows two paradigms: joint response-and-citation generation \cite{menick2022teaching, cohen2022lamda, glaese2022improving}, often restricted to single-source settings, and post-hoc attribution \cite{gao2023alce, yue2023automatic}, which struggles on human-annotated multi-source datasets. \cite{patel2024towards} addressed this with POLITICITE, a benchmark featuring multi-paragraph questions and human-labeled multi-source attributions.

Complementary to attribution methods, Inseq~\cite{sarti-etal-2023-inseq} provides a toolkit for token- and sequence-level attribution in neural text generation, enabling analysis of model behavior via gradient- and attention-based explanations. While Inseq is designed for textual inputs and does not natively support structured tables or reasoning-aligned cell-level grounding, we repurpose its attribution primitives as part of our framework to operate over table representations.

To improve traceability, \cite{khalifa2024sourceaware} proposed source-aware pretraining and instruction tuning by linking document IDs to citations, though adaptability to evolving knowledge remains limited. Attribution research has also expanded to multimodal QA, underscoring the need for explainable grounding across text, tables, and visual data. 

Existing methods lack fine-grained attribution in structured data that captures both final answers and intermediate reasoning , critical for transparent multi-hop inference. Current benchmarks omit reasoning-aligned cell-level annotations, and evaluation remains costly, subjective, or unsuitable for scale. We address these gaps by introducing a benchmark with manual cell-level attributions for both final and intermediate steps, a modular multi-agent framework for post-hoc cell-level attribution, and a scalable reference-free metric for attribution quality.

\section{\benchmarkName Benchmark}

\subsection{Limitations of the TabCite Benchmark}

TabCite~\cite{mathur2024matsa} is a closely related benchmark for table-based citation. While it enables coarse-grained evaluation, several aspects of its design limit its suitability for \emph{fine-grained} phrase-to-cell attribution.

\paragraph{Absence of phrase-cell alignment.}
In TabCite, attribution labels are defined at the level of the full answer, not individual answer phrases. As a result, the benchmark cannot verify whether each semantic unit in the answer is grounded in the correct cell.

For example, consider the answer \emph{“Wind Power, with an efficiency between 30\% and 45\%”} with attributed cells $\{[1,0],[1,2]\}$. Without phrase-level alignment, the evaluation cannot distinguish the correct mapping \emph{“Wind Power”} $\rightarrow [1,0]$, \emph{“30\%-45\%”} $\rightarrow [1,2]$ from the incorrect mapping where these associations are reversed. Both are treated as equally valid, even though only the first is semantically correct. This prevents TabCite from assessing fine-grained grounding.

\paragraph{Noise in ground-truth attributions.}
TabCite is constructed from ToTTo, FetaQA, and AITQA. Re-examining these sources reveals substantial label noise. In a manual inspection of 500 randomly sampled examples per dataset, we observe error rates of 21.7\% for \textit{ToTTo} and 55.3\% for \textit{FetaQA}. Moreover, \textit{AITQA} does not provide gold attribution labels, and its label construction procedure is undocumented.

\autoref{fig:mistake:FetaQA} shows a representative \textsc{FetaQA}:\\
\noindent\textbf{Question:} In which films did Pooja Ramachandran play the role of Cathy?\\
\noindent\textbf{Answer:} Pooja Ramachandran starred as Cathy in \textit{Kadhalil Sodhapuvadhu Yeppadi} and its Telugu version \textit{Love Failure}.

\medskip
While the relevant films and roles are correctly highlighted, the ground-truth attribution also marks several unnecessary cells, including other entries from the same year and an unrelated film (\textit{Pizza}).

\begin{table}[!ht]
\centering
\resizebox{\linewidth}{!}{%
\begin{tabular}{cccc}
\hline
\textbf{Year} & \textbf{Film} & \textbf{Role} & \textbf{Language} \\ \hline
2002 & Yathrakarude Sradakku & -- & Malayalam  \\ 
\cellcolor{incorrect}2012 & \cellcolor{correct}Kadhalil Sodhapuvadhu Yeppadi & \cellcolor{correct}Cathy & Tamil  \\ 
\cellcolor{incorrect}2012 & \cellcolor{correct}Love Failure & \cellcolor{correct}Cathy & \cellcolor{correct}Telugu  \\ 
2012 & Nanban & Jeeva's Wife & Tamil  \\ 
\cellcolor{incorrect}2012 & \cellcolor{incorrect}Pizza & Smitha & Tamil  \\ 
2013 & Swamy Ra Ra & Bhanu & Telugu  \\ \hline
\end{tabular}}
\caption{Example from \textsc{FetaQA} with ground-truth attribution highlighting (\textcolor{correct}{green} = relevant, \textcolor{incorrect}{red} = irrelevant).}
\label{fig:mistake:FetaQA}
\end{table}
\resizebox{0.95\linewidth}{!}{$
\{\textcolor{incorrect}{[2, 0]}, \textcolor{correct}{[2, 1]}, \textcolor{correct}{[2, 2]}, \textcolor{incorrect}{[3, 0]}, \textcolor{correct}{[3, 1]}, \textcolor{correct}{[3, 2]}, \textcolor{correct}{[3, 3]}, \textcolor{incorrect}{[5, 0]}, \textcolor{incorrect}{[5, 1]}\}
$}

These ground truth labels systematically over-include cells that are loosely related (e.g., same year, similar structure) rather than strictly necessary for the answer. This can encourage models to rely on partial or column-level matches instead of consistent row-level evidence.

Overall, the lack of phrase-level alignment and the presence of noisy or undocumented attributions limit TabCite’s usefulness for evaluating fine-grained table grounding. These limitations motivate \benchmarkName, which provides explicit phrase-to-cell alignment and carefully curated, transparent ground-truth annotations.

\subsection{\benchmarkName Construction}

To address the aforementioned limitations of TabCite~\citet{mathur2024matsa}, we construct a new benchmark by reformulating and manually annotating examples from three public datasets: ToTTo~\cite{parikh2020totto}, FetaQA~\cite{nan2021fetaqa}, and AITQA~\cite{katsis2022aitqa}.

\paragraph{ToTTo} is an open-domain, Wikipedia-based table-to-text dataset. Because it does not contain natural-language questions, we follow the approach of MATSA~\citet{mathur2024matsa} by treating the reference descriptions as answers and synthetically generating corresponding questions using DeepSeek-V3~\cite{deepseekai2025deepseekv3technicalreport}. ToTTo includes a wide range of table structures, such as merged cells, variable row counts, and nested headers, and thus captures much of the structural complexity found in real-world tables.

\paragraph{FetaQA} consists of free-form, multi-hop question-answer pairs over Wikipedia tables. Many questions require aggregating evidence from multiple cells across different rows. Although the tables typically lack deeply nested headers, the dataset presents substantial reasoning challenges due to its long-form answers and diverse table content.

\paragraph{AITQA} is a domain-specific question-answering dataset derived from SEC 10-K annual reports in the airline industry. Its tables contain complex column hierarchies, specialized financial terminology, and heterogeneous numerical formats, making the dataset representative of enterprise and scientific document settings.

\paragraph{Human Annotation.}  
From each dataset, we manually annotate 500 randomly sampled examples (all 513 examples from AITQA), yielding a gold set with precise, phrase-aligned cell-level attributions (\autoref{tab:dataset-stats}). For each example, every answer phrase is explicitly linked to the table cell(s) that support it, enabling fine-grained and interpretable evaluation. Inter-annotator agreement, measured using Cohen’s $\kappa$ on 250 randomly sampled instances across all three datasets, is 0.72, indicating substantial agreement.

\begin{table}[t]
\centering
\resizebox{\linewidth}{!}{%
\begin{tabular}{lccc}
\toprule
\textbf{Dataset} & \shortstack{\textbf{Total} \\ \textbf{Examples}} & \shortstack{\textbf{Gold Set} \\ \textbf{(Human-Annotated)}} & \shortstack{\textbf{Silver Set} \\ \textbf{(Original)}} \\
\midrule
ToTTo   & 7,700  & 500   & 7,200 \\
FetaQA  & 3,004  & 500   & 2,504 \\
AITQA   & 513    & 513   & --    \\
\bottomrule
\end{tabular}}
\caption{Dataset statistics for the \benchmarkName.}
\label{tab:dataset-stats}
\end{table}

The remaining ToTTo and FetaQA examples retain their original (and potentially noisy) attributions and form a silver set, which provides broader coverage for large-scale evaluation. Due to its smaller size, AITQA does not include a silver subset. Together, the gold and silver sets support both rigorous fine-grained evaluation and scalable benchmarking of attribution performance.

\section{\methodName Agentic Framework}

\begin{figure}[!ht]
    \centering
    \vspace{-1.0em}
    \includegraphics[width=\linewidth]{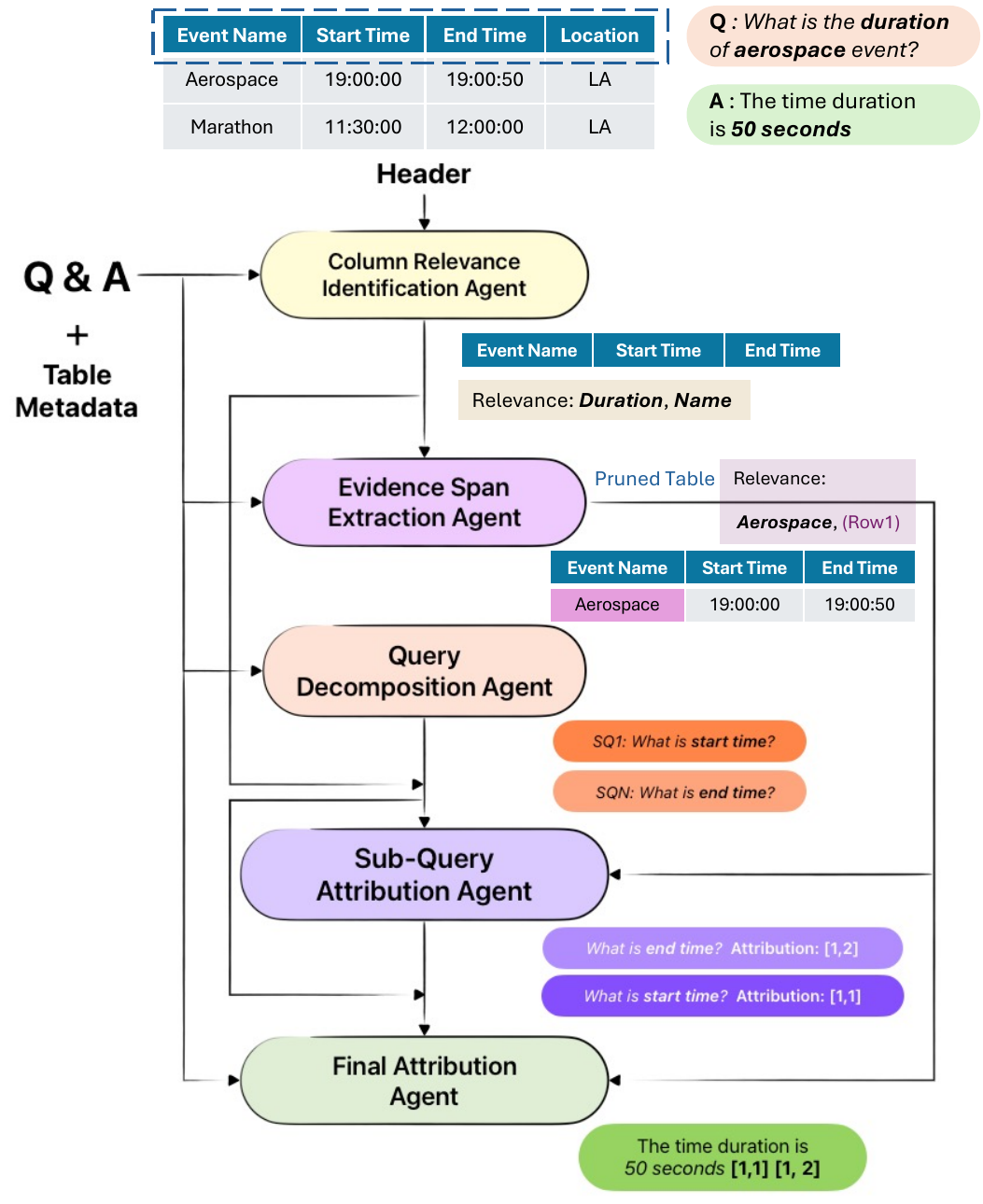}
    \vspace{-1.0em}
    \caption{Architecture diagram of the \methodName.}
    \label{fig:methodology}
\end{figure}

\subsection{Task Formulation}

We consider \emph{cell-level structured single-table attribution}, where the goal is to identify the minimal yet sufficient set of table cells that support a model-generated answer.

Let $T$ be a table with $R$ rows and $C$ columns, and let each cell be indexed as $T[i,j]$ for $i \in [1,R]$ and $j \in [1,C]$. 
Given a natural-language question $q$ and its answer $a$, we seek an attribution set $\mathcal{A}_T \subseteq \{T[i,j]\}$ such that $
f(T, q, a) \rightarrow \mathcal{A}_T$.

The set $\mathcal{A}_T$ must (i) be sufficient and non-contradictory for deriving $a$ from $T$, and (ii) ground all semantic content in $a$.
It may include both \emph{explicit cells}, whose values appear verbatim in $a$, and \emph{implicit cells}, which are required for intermediate reasoning but not directly mentioned.
As illustrated in \autoref{fig:introduction}, correctly answering the efficiency question for \textit{Wind Power} requires cost and scalability cells in addition to the final efficiency cell.

\subsection{\methodName Framework}

To solve this task, we propose \methodName, an LLM-based multi-agent framework in which each agent handles a distinct subtask in the attribution pipeline.
The framework is designed for structured and semi-structured single tables and scales to arbitrary table size and schema.
\autoref{fig:methodology} gives an overview.

\paragraph{1. Column Relevance Identification.}
This agent selects the subset of columns needed to derive the answer. 
Relevant columns may be \emph{explicit}, whose values appear in the answer, or \emph{implicit}, required for intermediate computations.

Given table metadata and schema, together with $(q,a)$, the agent uses few-shot prompting to predict the full set of relevant columns (\autoref{fig:methodology}). 
For example, for the question \emph{“What is the duration of the event?”} with answer \emph{“50 seconds”}, it must identify \textit{Start\_Time} and \textit{End\_Time} even though no column is labeled “duration.”  
Similarly, in \autoref{fig:introduction}, answering the constrained efficiency query requires \textit{Cost}, \textit{Scalability}, and \textit{Efficiency}, even though only the efficiency value appears in the final answer.

\paragraph{2. Evidence Span Extractor.}
Given the selected columns, this agent identifies the subset of rows required to derive the answer.
Following the SQL-based row extraction strategy of \citet{abhyankar2025hstarllmdrivenhybridsqltext}, it uses few-shot prompting to generate filtering conditions over the relevant columns and metadata.

In \autoref{fig:methodology}, for the question \emph{“What is the duration of the aerospace event?”}, the agent can generate a filter such as \texttt{WHERE Event\_Name = 'Aerospace'}, retaining only the row for the Aerospace event.
In \autoref{fig:introduction}, more complex multi-step filtering is required: rows are filtered by \textit{Cost} $\leq 50$/MWh, then by \textit{Scalability} $\geq 3$, and finally ranked by \textit{Efficiency}.
This pruning reduces computational overhead while preserving the complete reasoning chain.

\paragraph{3. Query Decomposition.}
This agent decomposes the original question into semantically coherent sub-questions, each corresponding to an intermediate reasoning step needed to reconstruct the answer.
The goal is to surface implicit dependencies that holistic attribution often misses.

Given table metadata, $(q,a)$, and the pruned table, the agent generates sub-questions whose answers collectively entail the original response (\autoref{fig:methodology}).
For example, \emph{“What is the duration of the event Marathon?”} may be decomposed into \emph{“What is the start time of the event?”} and \emph{“What is the end time of the event?”}, explicitly exposing the role of \textit{Start\_Time} and \textit{End\_Time}.
To ensure faithfulness, each generated fact is checked by a pretrained NLI model (RoBERTa; \citealt{liu2019robertarobustlyoptimizedbert}), following \citet{mathur2024matsa}.

In \autoref{fig:introduction}, the comparative efficiency question is similarly decomposed into sub-questions for cost filtering, scalability filtering, and efficiency selection, revealing how multiple columns contribute to the final answer.

\paragraph{4. Sub-Query Attribution.}
This agent performs fine-grained attribution by grounding each sub-question to its supporting cells.
Given the sub-questions and the pruned table, it outputs cell coordinates $(i,j)$ corresponding to the evidence used to answer each sub-query.

In \autoref{fig:methodology}, sub-questions about start and end time are mapped to the \textit{Start\_Time} and \textit{End\_Time} cells in the relevant row; these cells are essential for computing the duration, even though they are absent from the answer text.
In \autoref{fig:introduction}, sub-questions are similarly grounded to the \textit{Cost}, \textit{Scalability}, and \textit{Efficiency} cells that jointly determine the answer.
Operating at the cell level allows the agent to capture such implicit evidence, which row- or column-level attribution would miss.

\paragraph{5. Final Attribution.}
The final agent consolidates all intermediate attributions into a single phrase-to-cell mapping.
It aligns the cell coordinates produced by the Sub-Query Attribution agent with spans in the original answer, yielding a complete, interpretable grounding of each answer component in verifiable tabular evidence and enabling faithful evaluation of table-based QA systems.

\section{Experimental Setup}

\paragraph{Baselines.}
We evaluate our approach \benchmarkName\ and compare against the following baselines. 
Unless otherwise noted, all methods use \texttt{GPT-4o}~\cite{openai2024gpt4ocard} as the LLM. Finally, all prompt of our method are provided in Appendix~\ref{supsec:prompts}

\textbf{Few-shot ICL}~\cite{gao2023enablinglargelanguagemodels} prompts the model with a small number of annotated examples and directly asks it to produce cell-level attributions for each question-answer pair.  

\textbf{Post-hoc Retrieval}~\cite{gao-etal-2023-rarr} linearizes each table row into text and uses a dense retriever (Sentence-BERT~\cite{reimers2019sentencebertsentenceembeddingsusing}) to rank rows by cosine similarity to the concatenated question and answer; the top-$k$ rows ($k{=}10$) are then passed to the LLM to generate cell-level attributions.  

\textbf{\textsc{GeneratePrograms}}~\cite{wan2025generationprogramsfinegrainedattributionexecutable} converts each row into atomic fact sentences, assigns each cell a unique identifier, and applies the original \textsc{GeneratePrograms} pipeline unchanged to produce executable programs and their induced attributions.  

\textbf{\textsc{Inseq}-based Attribution}~\cite{sarti-etal-2023-inseq} repurposes token- and sequence-level attribution by linearizing tables into structured text and treating cell values as attribution units; gradient- and attention-based scores are aggregated at the cell level. 
For this baseline we use \texttt{Qwen3-30B-A3B}~\cite{yang2025qwen3technicalreport}, as gradient-based attribution requires direct access to model parameters.

To our knowledge, no existing method natively supports reasoning-aligned cell-level attribution over structured tables. 
The closest framework, MATSA~\cite{mathur2024matsa}, operates at row-column granularity and is not publicly available, precluding direct comparison.

\paragraph{Evaluation Metrics.}
We evaluate attribution quality at three levels of granularity: \textbf{row}, \textbf{column}, and \textbf{cell} using precision and recall computed from the overlap between predicted and gold-standard attribution sets.

Let $R'$, $C'$, and $S'$ denote the sets of predicted rows, columns, and cells, respectively, and let $\hat{R}$, $\hat{C}$, and $\hat{S}$ denote the corresponding gold-standard sets. Precision measures the proportion of predicted elements that are correct, while recall measures the proportion of gold elements that are successfully retrieved.

\begin{table}[!htbp]
\centering
\small
\renewcommand{\arraystretch}{1.15}
\begin{tabular}{lcc}
\toprule
\textbf{Granularity} & \textbf{Precision} & \textbf{Recall} \\
\midrule
Row    & $\displaystyle \frac{|R' \cap \hat{R}|}{|R'|}$ & $\displaystyle \frac{|R' \cap \hat{R}|}{|\hat{R}|}$ \\
Column & $\displaystyle \frac{|C' \cap \hat{C}|}{|C'|}$ & $\displaystyle \frac{|C' \cap \hat{C}|}{|\hat{C}|}$ \\
Cell   & $\displaystyle \frac{|S' \cap \hat{S}|}{|S'|}$ & $\displaystyle \frac{|S' \cap \hat{S}|}{|\hat{S}|}$ \\
\bottomrule
\end{tabular}
\caption{Precision and Recall at different granularity.}
\label{tab:eval-metrics}
\end{table}

Evaluating across multiple granularity enables analysis of attribution performance at both coarse and fine levels: row- and column-level metrics capture broader localization accuracy, while cell-level metrics assess fine-grained grounding fidelity.

\begin{table*}[!ht]
\centering
\begin{tabular}{lcccccc}
\hline \hline
                                  & \multicolumn{2}{c}{\textbf{ToTTo}} & \multicolumn{2}{c}{\textbf{FetaQA}} & \multicolumn{2}{c}{\textbf{AITQA}} \\ \cline{2-7} 
\multirow{-2}{*}{\textbf{Method}} & Precision        & Recall          & Precision        & Recall           & Precision        & Recall          \\
\multicolumn{7}{c}{\cellcolor[HTML]{E6FFE6}\textit{Row-Level Attribution}}                                               \\
SBERT + \texttt{GPT-4o}     & 43.38          & 39.09          & 57.02          & 55.40          & 66.82          & 68.10          \\
\textsc{GenerationPrograms} & 50.00          & 31.28          & 75.00          & 40.06          & 59.68          & 71.90          \\
Fewshot + CoT      & 17.30          & 12.80          & 34.40          & 30.10          & 04.30          & 04.30          \\
\textsc{Inseq}              & 37.50               & {\ul 74.50}               & 56.40	                & {\ul 84.70}                & 31.20	 & \textbf{97.60}               \\
\methodName{} - Lite                  & {\ul 77.00}      & 62.60     & {\ul 83.00}      & 79.20      & {\ul 91.30}      & 92.90     \\
\methodName{}                         & \textbf{71.19}   & \textbf{80.38}  & \textbf{94.30}   & \textbf{93.36}   & \textbf{96.65}   & {\ul 97.12}  \\

\multicolumn{7}{c}{\cellcolor[HTML]{DEF6FC}\textit{Column-Level Attribution}}                                            \\
SBERT + \texttt{GPT-4o}     & 90.51    & \textbf{85.91} & 94.67          & \textbf{84.77} & 46.68          & 97.14    \\
\textsc{GenerationPrograms} & 71.81          & 24.71          & 78.42          & 20.00          & 49.86          & 83.81          \\
Fewshot + CoT      & \textbf{92.70} & 76.40          & 95.80          & 67.30          & 47.50          & 94.30          \\

\textsc{Inseq}              & 73.10  &  74.10    & 82.60	  & 65.50   & 34.70	 &  \textbf{99.98}              \\

\methodName{} - Lite   & 88.60          & 48.90          & {\ul 94.70}    & 54.80          & \textbf{79.20} & 85.30          \\
\methodName{}         & {\ul 91.50}          & {\ul 77.64}    & \textbf{96.39} & {\ul 83.07}    & {\ul 54.09}    & {\ul 98.09} \\

\multicolumn{7}{c}{\cellcolor[HTML]{FAE4CA}\textit{Cell-Level Attribution}}                                              \\
SBERT + \texttt{GPT-4o}     & 39.78          & 36.97          & 52.08          & 46.16          & 31.96          & 66.67          \\
\textsc{GenerationPrograms} & 29.35          & 13.61          & 50.78          & 15.74          & 30.32          & 67.14          \\
Fewshot + CoT      & 14.50          & 10.10          & 27.40          & 17.80          & 02.20          & 04.30          \\

\textsc{Inseq}              & 42.70	 & {\ul 53.80}               &  56.50	              & {\ul 44.20}               &  19.20	              &  \textbf{97.10}              \\

\methodName{} - Lite                  & {\ul 73.80}      & 39.60     & {\ul 75.40}      & 42.30      & \textbf{73.70}   & 80.60     \\
\methodName{}                         & \textbf{74.20}   & \textbf{67.05}  & \textbf{89.81}   & \textbf{78.84}   & {\ul 52.37}      & {\ul 95.22}  \\ \hline \hline
\end{tabular}
\caption{\textbf{Attribution performance across granularities and datasets.} 
Precision and recall for row-, column-, and cell-level attribution on ToTTo, FetaQA, and AITQA. 
Blocks correspond to different granularities. For each dataset-granularity pair, \textbf{bold} best method; {\ul underline} marks second best}

\label{tab:attribution-results}
\end{table*}

\subsection{Main Results}
\label{sec:main-results}

\autoref{tab:attribution-results} reports precision and recall for row-, column-, and cell-level attribution on ToTTo, FetaQA, and AITQA. We highlight the main trends here and refer to the table for complete figures.

\paragraph{Row-Level Attribution.}
\methodName{} achieves the best row-level performance on all three datasets.
On ToTTo, it improves precision from a best baseline of 50.00 (\textsc{GenerationPrograms}) to 71.19 and recall from 74.50 (\textsc{Inseq}) to 80.38.
On FetaQA, \methodName{} attains 94.30 precision and 93.36 recall, a gain of roughly 19 precision points over the strongest baseline (75.00) and +8.7 recall over the best baseline (84.70).
On AITQA, it reaches 96.65 precision versus 66.82 for the best baseline and matches the high-recall regime (97.12 vs.\ 97.60 for \textsc{Inseq}), showing that schema-aware filtering avoids the over-retrieval that hurts baseline precision.

\paragraph{Column-Level Attribution.}
Column attribution is strong for several methods, but \methodName{} is consistently competitive or best.
On ToTTo, it nearly matches the highest baseline precision (91.50 vs.\ 92.70) while maintaining solid recall (77.64).
On FetaQA, it achieves both the highest precision (96.39) and near-best recall (83.07 vs.\ 84.77 for SBERT+\texttt{GPT-4o}).
On AITQA, \methodName{} attains 54.09 precision and 98.09 recall, improving precision over the best baseline (49.86) while staying in the same high-recall regime as SBERT+\texttt{GPT-4o} and \textsc{Inseq}.

\paragraph{Cell-Level Attribution.}
Cell-level attribution is the hardest setting, and here \methodName{} shows the largest margins.
On ToTTo, it raises precision from 42.70 and recall from 53.80 (both from \textsc{Inseq}) to 74.20 and 67.05.
On FetaQA, it improves precision from 56.50 to 89.81 and recall from 46.16 to 78.84, a gain of more than 30 points on both metrics.
On AITQA, \methodName{} boosts precision from 31.96 (SBERT+\texttt{GPT-4o}) to 52.37 while retaining very high recall (95.22 vs.\ 97.10 for \textsc{Inseq}).
These gains indicate that query decomposition and sub-query attribution are particularly effective at isolating the exact supporting cells, including implicit evidence.

\paragraph{Baseline Behavior and Ablations.}
Few-shot CoT generally underperforms, especially at the cell level, confirming that unguided prompting lacks the structured reasoning needed for fine-grained grounding.
SBERT+\texttt{GPT-4o} and \textsc{Inseq} often achieve strong recall (e.g., row-level recall of 97.60 on AITQA), but their precision remains low due to aggressive retrieval and limited schema awareness.
\textsc{GenerationPrograms} trades precision for recall without closing this gap.
\methodName{}-Lite (uses \texttt{Qwen2.5-7B-Instruct}~\cite{qwen2025qwen25technicalreport} as LLM), which omits some components of our full framework, already outperforms all baselines in most settings, while the full \methodName{} consistently delivers the best precision-recall balance across datasets and granularities.

\subsection{Variants and Ablations}
\label{sec:variants-ablation}

\autoref{tab:ablation} reports cell-level precision and recall for variants of \methodName{} and ablations of its main components.

\paragraph{Variants.}
Both pipeline variants underperform the full system. 
Running \emph{query decomposition before table pruning} yields lower precision on all datasets (e.g., 63.00 vs.\ 74.20 on ToTTo), suggesting that pruning benefits from operating on the original table rather than on decomposed sub-questions.  
Processing \emph{one subquery at a time} further reduces precision (61.32 on ToTTo and 69.67 on FetaQA), indicating that joint reasoning over sub-questions is important for maintaining consistent evidence selection.  
The full \methodName{} achieves the best trade-off, with 74.20/67.05 on ToTTo and 89.81/78.84 on FetaQA.

\paragraph{Ablations.}
Removing \emph{table pruning} slightly improves precision on AITQA (86.33 vs.\ 52.37) but reduces precision on ToTTo and FetaQA and lowers recall overall, confirming that pruning is generally beneficial for accuracy and efficiency.  
The largest degradation occurs when removing \emph{query decomposition}: precision drops to 56.00 on ToTTo and 65.40 on FetaQA, with recall also reduced.  
This shows that decomposition is crucial for surfacing intermediate evidence and aligning attributions with multi-step reasoning.

Overall, these results indicate that both table pruning and query decomposition are necessary: pruning controls noise, while joint reasoning over decomposed sub-questions enables accurate cell-level grounding.

\begin{table*}[!htbp]
\centering
\resizebox{\linewidth}{!}{%
\begin{tabular}{lcccccc}
\toprule
\textbf{Method} & \multicolumn{2}{c}{\textbf{ToTTo}} & \multicolumn{2}{c}{\textbf{FetaQA}} & \multicolumn{2}{c}{\textbf{AITQA}} \\
\cmidrule(lr){2-3} \cmidrule(lr){4-5} \cmidrule(lr){6-7}
& Precision & Recall & Precision & Recall & Precision & Recall \\
\midrule
\multicolumn{7}{c}{\textit{Variants of ~\methodName}} \\
Query Decomposition before Table Pruning & 63.00 & 60.15 & 71.23 & 75.10 & 85.38 & 92.89 \\
Passing One Subquery at a Time & 61.32 & 60.10 & 69.67 & 73.33 & \textbf{85.95} & 92.00 \\
 \methodName{}  & \textbf{74.20} & \textbf{67.05} & \textbf{89.81} & \textbf{78.84} & 52.37 & \textbf{95.22} \\
\midrule
\multicolumn{7}{c}{\textit{Ablation Study on ~\methodName}} \\
 \methodName{}  & \textbf{74.20} & \textbf{67.05} & \textbf{89.81} & \textbf{78.84} & 52.37 & \textbf{95.22} \\
 - w/o Table Pruning & 73.14 & 60.10 & 72.89 & 75.78 & \textbf{86.33} & 93.10 \\
 - w/o Query Decomposition & 56.00 & 56.30 & 65.40 & 68.32 & 47.22 & 91.80 \\
\bottomrule
\end{tabular}}
\caption{\textbf{Variants and ablations of \methodName{} for cell-level attribution.}
Cell-level precision and recall on ToTTo, FetaQA, and AITQA for (top) pipeline variants that modify how sub-questions are processed and (bottom) ablations that remove specific modules. \textbf{Bold} indicate the best score.}

\label{tab:ablation}
\end{table*}

\begin{figure*}[!ht]
    \centering
    \includegraphics[width=\linewidth]{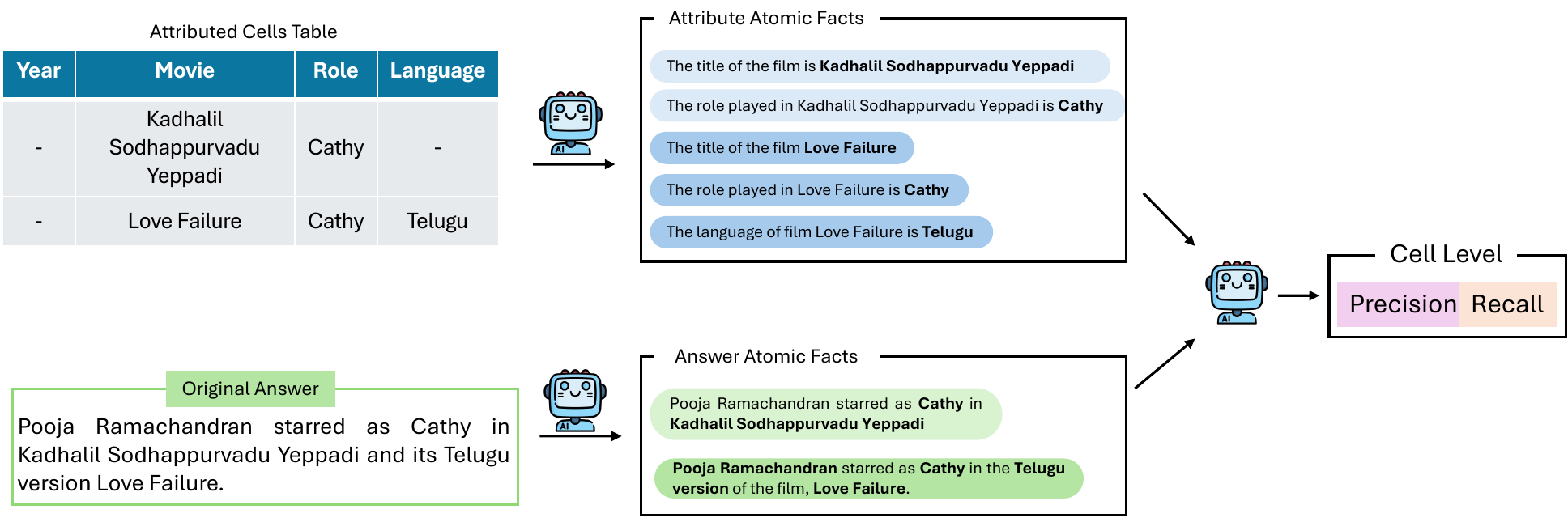}
    \vspace{-1.0em}
    \caption{Illustration of Reference-less Evaluation Metric \metricName}
    \label{fig:Ref-less}
    \vspace{-1.0em}
\end{figure*}

\begin{figure}[!ht]
    \centering
    \includegraphics[width=\linewidth]{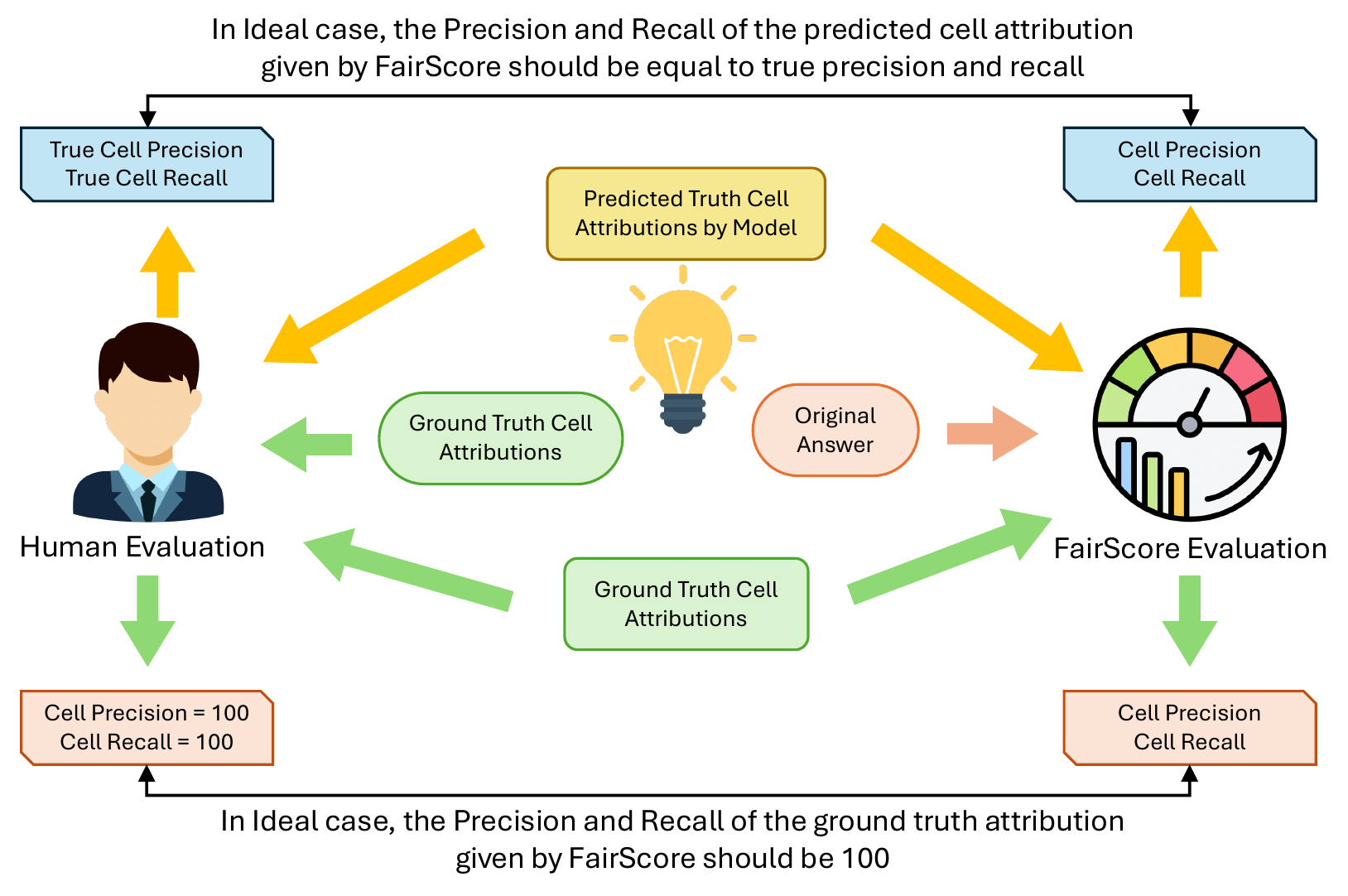}
    \vspace{-1.0em}
    \caption{Systematic analysis of the metric -  \metricName }
    \label{fig:eval_metric_analysis}
    \vspace{-0.55em}
\end{figure}

\section{\metricName Metric}

To reduce the cost of manually annotating cell-level attributions, we introduce \metricName, a \emph{reference-less} evaluation pipeline that estimates attribution quality by comparing \emph{atomic facts} derived from predicted cells with atomic facts extracted from the answer (\autoref{fig:Ref-less}). The pipeline has three stages.

\begin{enumerate}
    \item \textbf{Cell-to-fact conversion.}  
    Each cell predicted as relevant is converted into a short, declarative \emph{atomic fact} by combining its value with the corresponding column header and, when needed, a row key.  
    For example, the cells (\textit{Kadhalil Sodhapuvadhu Yeppadi}, \textit{Cathy}) under columns (\textit{Film}, \textit{Role}) yield facts such as  
    \emph{``The role of Cathy appears in the film Kadhalil Sodhapuvadhu Yeppadi.''}  
    This step turns structured evidence into natural-language statements that can be compared to answer-derived facts.

    \item \textbf{Answer-to-fact conversion.}  
    The reference answer is decomposed into minimal, self-contained atomic facts using an LLM.  
    For instance, the answer  
    \emph{``Pooja Ramachandran starred as Cathy in Kadhalil Sodhapuvadhu Yeppadi and its Telugu version Love Failure.''}  
    can be split into  
    \emph{``Pooja Ramachandran starred as Cathy in Kadhalil Sodhapuvadhu Yeppadi.''} and  
    \emph{``Pooja Ramachandran starred as Cathy in the Telugu version Love Failure.''}  
    This yields fact set against which table-derived facts can be aligned.

    \item \textbf{Fact alignment and comparison.}  
    A scoring LLM compares the two fact sets and decides, for each pair, whether the cell-derived fact correctly supports an answer fact.  
    Unmatched answer facts correspond to \emph{recall errors} (missing attributions), whereas unmatched cell-derived facts correspond to \emph{precision errors} (over-attribution).  
    In this way, we can evaluate attribution quality without access to gold cell labels.
\end{enumerate}

\paragraph{\metricName Precision and Recall.}
Let $a$ be the number of answer-derived facts that are supported by at least one predicted cell-derived fact, and $b$ the number of unsupported answer facts.  
Let $c$ be the number of predicted cell-derived facts that correctly match an answer fact, and $d$ the number of predicted facts that do not match any answer fact.  
We define
\begin{equation*}
\text{Recall} = \frac{a}{a + b}, \qquad
\text{Precision} = \frac{c}{c + d}.
\end{equation*}
Recall (answer$\rightarrow$cell view) measures how completely the predicted attributions cover the answer content, while precision (cell$\rightarrow$answer view) measures how many predicted evidential cells are actually needed for the answer.  
Together, these quantities provide an interpretable, scalable proxy for cell-level attribution quality.

\subsection{\metricName Systematic Analysis}
We next assess the reliability of \metricName\ as a surrogate for human evaluation (\autoref{fig:eval_metric_analysis}).  
Our analysis considers two complementary aspects:  
(i) how closely \metricName\ precision and recall track true cell-level precision and recall when gold attributions are available, and  
(ii) how consistently \metricName\ scores distinguish correct from incorrect attributions when applied both to model predictions and to gold cell sets themselves.

\paragraph{1. Human agreement check.}
We first apply \metricName\ to the 1{,}500 gold examples, pairing human cell-level attributions with their reference answers. 
Since these labels are manually annotated, we expect scores close to 1.0. 
In this setting, \metricName\ attains 83.20 precision and 88.34 recall, indicating strong alignment with human judgments. 
The remaining gap is largely due to variability in LLM-generated atomic facts and occasional errors in the entailment-based alignment.

\paragraph{2. Alignment with attribution evaluation.}
We next assess how well \metricName\ approximates true attribution quality when gold labels are used only for analysis. \autoref{tab:fairscore_delta_cellonly} compares \metricName-predicted precision and recall with human-annotated cell-level scores for \methodName{} on the same 1{,}500 examples.

\begin{table}[!ht]
\centering
\resizebox{\linewidth}{!}{%
\begin{tabular}{lccc ccc}
\toprule
\textbf{Dataset} & \textbf{Pred P} & \textbf{Actual P} & $\Delta$P & \textbf{Pred R} & \textbf{Actual R} & $\Delta$R \\
\midrule
ToTTo   & 58.81 & 74.52 & -15.71 & 56.36 & 67.31 & -10.95 \\
FetaQA  & 73.45 & 89.81 & -16.36 & 72.40 & 78.84 & -06.44 \\
AITQA   & 48.10 & 52.38 & -04.28 & 57.27 & 95.65 & -38.38 \\
\bottomrule
\end{tabular}}
\caption{\metricName-predicted vs.\ human-annotated cell-level precision and recall for \methodName's attributions. $\Delta$ columns show prediction error.}
\label{tab:fairscore_delta_cellonly}
\end{table}

\noindent\textbf{Analysis} across datasets, \metricName\ systematically underestimates absolute scores but tracks relative trends. 
On ToTTo and FetaQA, precision is underestimated by about 16 points and recall by 6-11 points. 
AITQA exhibits a larger recall gap ($-38.38$), reflecting the greater linguistic and structural variability of financial tables, while precision remains within 4.3 points. 
Despite these discrepancies, \metricName\ preserves the correct ranking and relative differences, supporting its use as a scalable proxy for human cell-level evaluation.

Taken together, these results indicate that \metricName\ closely mirrors human judgments of attribution quality. 
By converting predicted cells and answers into atomic facts and aligning them via entailment-based matching, it captures both explicit and implicit evidence while penalizing omissions and hallucinations, enabling robust and scalable evaluation without additional gold labels.

\subsection{Evaluation with \metricName}

\begin{table}[!ht]
\small
\centering
\resizebox{\linewidth}{!}{%
\begin{tabular}{lcccccc}
\toprule
\textbf{Method} & \multicolumn{2}{c}{\textbf{ToTTo}} & \multicolumn{2}{c}{\textbf{FetaQA}} & \multicolumn{2}{c}{\textbf{AITQA}} \\
\cmidrule(lr){2-3} \cmidrule(lr){4-5} \cmidrule(lr){6-7}
                       & P & R & P & R & P & R \\
\midrule
Fewshot + CoT           & 30.81 & 13.25 & 15.67 & 17.73 & 11.73 &  6.69 \\
SBERT + \texttt{GPT-4o} & 20.51 & 16.84 & 20.05 & 21.87 &  4.51 &  5.47 \\
\textsc{Inseq}         & 16.85 & 18.95 & 15.48 & 17.94 & 10.99 & 16.53 \\
\textsc{GP} & 14.96 & 11.32 & 16.04 & 14.13 &  7.62 &  3.77 \\
\methodName-Lite            & 53.87 &	40.20 &	51.88 &	45.12 &	\textbf{63.44} &	\textbf{55.13} \\
\methodName            & \textbf{56.89} & \textbf{48.39} & \textbf{63.73} & \textbf{64.15} & 42.20 & 49.93 \\
\bottomrule
\end{tabular}}
\caption{Cell-level precision and recall predicted by \metricName\ for different attribution methods on the gold sets of ToTTo, FetaQA, and AITQA. Note: here P refers to Precison and R refers to recall, Method \textsc{GP} refers to \textsc{GenerationPrograms}.}
\label{tab:FairScore eval}
\end{table}

We report \metricName\ scores in terms of precision (P) and recall (R).

\paragraph{Methods evaluated with \metricName\ on gold sets.}
\autoref{tab:FairScore eval} shows \metricName\ scores for \methodName, \methodName-Lite, and all baselines on the 1{,}500 gold examples. 
On \textbf{ToTTo}, \methodName\ reaches P/R of 56.89/48.39, improving over the strongest baseline (Few-shot ICL: 30.81 P; \textsc{Inseq}: 18.95 R) by more than 25 points in P and nearly 30 in R.  
On \textbf{FetaQA}, it achieves 63.73/64.15, exceeding the best baseline (SBERT+\texttt{GPT-4o}: 20.05/21.87) by over 40 points on both metrics.  
On \textbf{AITQA}, \methodName-Lite attains the highest scores (63.44/55.13), while the full \methodName\ still substantially outperforms all non-\methodName\ variants (42.20/49.93).  
Overall, \metricName\ clearly separates strong from weak attribution methods and preserves the performance ordering observed with gold-label evaluation.

\paragraph{\methodName\ evaluated with \metricName\ on silver sets.}
We also apply \metricName\ to silver sets instances without human annotations from ToTTo (7{,}200 examples) and FetaQA (2{,}504 examples).  
On ToTTo, \methodName\ attains 55.10/76.12; on FetaQA, it reaches 69.30/68.93. These scores are consistent with the corresponding gold-set estimates (56.89/48.39 for ToTTo and 63.73/64.15 for FetaQA) and preserve relative trends across datasets. Taken together, these results indicate that \metricName\ is a reliable reference-less proxy for human evaluation, enabling large-scale benchmarking of cell-level attribution when gold labels are scarce.

\section{Conclusion and Future Work}

We presented \methodName, a modular multi-agent framework for fine-grained cell-level attribution in table QA, and \benchmarkName, a 1{,}500-example benchmark with phrase-to-cell annotations drawn from ToTTo, FetaQA, and AITQA. 
We also introduced \metricName, a reference-less metric based on atomic fact alignment that enables scalable evaluation without gold cell labels. 
Experiments show that \methodName\ consistently outperforms strong baselines across datasets and granularities, and that \metricName\ closely tracks human judgments while preserving relative performance trends.

Future work includes extending \methodName\ to more complex reasoning (e.g., aggregation and multi-hop over multiple tables) and improving \metricName’s robustness so it can be applied more broadly across domains and modalities.

% \section*{Limitations}

% While \methodName provides interpretable, fine-grained cell-level attribution, it has several inherent limitations. First, it currently operates only on single-table settings and does not support multi-table joins, hierarchical schemas, or semi-/multimodal table structures, limiting applicability to more complex real-world QA scenarios. Second, the accuracy of core modules, particularly sub-question decomposition and cell attribution, relies heavily on large language models (LLMs). This dependence introduces sensitivity to prompt phrasing, model updates, and domain shifts, and results may vary substantially across providers and contexts. Third, the reference-less evaluation metric, \metricName, while scalable, depends on the quality of fact extraction and alignment and exhibits a systematic underestimation of absolute precision, particularly for compositional or ambiguous answers, and its robustness under domain shift remains untested. Fourth, the comparative scope is constrained by the unavailability of certain close baselines (e.g., MATSA), limiting direct head-to-head performance validation against the most relevant prior art. Finally, \benchmarkName covers only English-language tables from Wikipedia-like sources, and its generalizability to domain-shifted, multilingual, or more heterogeneous data distributions remains untested.

\section*{Limitations}

\methodName has several limitations. First, it is restricted to single-table settings and does not yet support multi-table joins, hierarchical schemas, or semi-/multimodal tables, limiting applicability to more complex QA scenarios. Second, its core components (e.g., query decomposition and cell attribution) rely heavily on LLMs, making performance sensitive to prompt design, model changes, and domain shift. Third, while \metricName\ is scalable, it depends on the quality of fact extraction and entailment-based alignment and tends to underestimate absolute precision, especially for compositional or ambiguous answers. Fourth, our comparisons are constrained by the lack of publicly available implementations for some close baselines (e.g., MATSA). Finally, \benchmarkName\ currently covers only English Wikipedia-style tables, so generalization to multilingual or domain-specific and more heterogeneous settings remains untested.

\bibliography{custom}
\bibliographystyle{acl_natbib}

\onecolumn

\appendix

\section*{Appendix}

\section{Prompt Templates}
\label{supsec:prompts}

\mytcbinput{prompts/Column_Relevance_Agent.tex}{Column Relevance Identification Agent}{0}{bw:domain}{prompt:column_re}
\mytcbinputwide{prompts/Evidence_Span_Extractor_Agent.tex}{Evidence Span Extractor Agent}{0}{bw:domain}{prompt:evidence_span}
\mytcbinputwide{prompts/Query_decomposition_Agent.tex}{Query Decomposition Agent}{0}{bw:domain}{prompt:query_decomp}
\mytcbinputwide{prompts/Sub-Query_Attribution_Agent.tex}{Sub-Query Attribution Agent}{0}{bw:domain}{prompt:sub_q_att}
\mytcbinputwide{prompts/Final_Attribution_Agent.tex}{Final Attribution Agent}{0}{bw:domain}{prompt:final}

\section{Ethics Statement}

We affirm that this work adheres to established ethical standards in NLP research and publication. All datasets used in our study ToTTo, FetaQA, and AITQA are publicly available and used in compliance with their respective terms. Our benchmark, \benchmarkName, is constructed through manual annotation with care to ensure fairness, factual accuracy, and privacy. We disclose all model configurations, prompting strategies, and evaluation protocols to support transparency and reproducibility. While \metricName is designed to eliminate the need for human labels, it depends on LLM judgments, which may reflect inherent biases; we mitigate this by using temperature-controlled generation and standardized prompts. AI assistance was used in experiments and drafting to improve clarity, structure, and analysis quality, with human oversight throughout.

\end{document}